\documentclass[sigconf]{acmart}

\AtBeginDocument{%
  \providecommand\BibTeX{{%
    \normalfont B\kern-0.5em{\scshape i\kern-0.25em b}\kern-0.8em\TeX}}}

\setcopyright{acmcopyright}
\copyrightyear{2023}
\acmYear{2023}
\acmDOI{10.1145/1122445.1122456}

\usepackage{comment}
\usepackage{enumerate}
\usepackage[shortlabels]{enumitem}
\usepackage[singlelinecheck=false,justification=justified]{caption}
\usepackage{algorithm} 
\usepackage{algpseudocode} 
\usepackage{amsmath}
\usepackage{multirow}
\usepackage{float}
\usepackage{lipsum}

\usepackage{graphics}
\usepackage{tikz,pgfplots}
\usepackage{soul}
\usepackage{capt-of}
\usepackage{caption}
\usepackage{soul}
\usepackage{makecell}
\usepackage{bm}
\usepackage{booktabs} 
\usepgfplotslibrary{groupplots}

\usepackage{amsthm}
\usepackage{etoolbox}

\AfterEndEnvironment{defn}{\noindent\ignorespaces}

\usepackage{booktabs,caption}
\usepackage[flushleft]{threeparttable}

\usepackage{kotex}

\newcommand{\todoblue}[1]{\textcolor{blue}{#1}}

\definecolor{customblue}{HTML}{20639B}
\definecolor{customred}{HTML}{ED553B}
\definecolor{customyellow}{HTML}{EB9605}
\definecolor{custompurple}{HTML}{9867C5}
\definecolor{customgreen}{HTML}{35B37E}

\pgfplotsset{compat=1.18}
\begin{document}

\title{ST-RAP: A Spatio-Temporal Framework for Real Estate Appraisal}

\settopmatter{authorsperrow=5}
\author{Hojoon Lee}
\authornote{Equal Contribution. Order determined by rolling dice.}
\email{joonleesky@kaist.ac.kr}
\affiliation{%
  \institution{KAIST AI}
}

\author{Hawon Jeong}
\authornotemark[1]
\email{hawon@kaist.ac.kr}
\affiliation{
  \institution{KAIST AI}
} 

\author{Byungkun Lee}
\authornotemark[1]
\email{byungkun@kaist.ac.kr}
\affiliation{%
  \institution{KAIST AI}
} 

\author{Kyungyup Lee}
\email{lkylove2323@gmail.com}
\affiliation{%
  \institution{Spacewalk}
} 

\author{Jaegul Choo}
\email{jchoo@kaist.ac.kr}
\affiliation{
  \institution{KAIST AI}
} 

\renewcommand{\shortauthors}{Lee et al.}

\ccsdesc[500]{Information systems~Spatial-temporal systems}

\keywords{Real Estate Appraisal; Spatio-Temporal Network; Graph Network} 




\newcommand{\ourmodel}{our model }
\newcommand{\alevel}{acceptance level }
\newcommand{\rlevel}{rejection level}

\begin{abstract}

In this paper, we introduce ST-RAP, a novel Spatio-Temporal framework for Real estate APpraisal. ST-RAP employs a hierarchical architecture with a heterogeneous graph neural network to encapsulate temporal dynamics and spatial relationships simultaneously. Through comprehensive experiments on a large-scale real estate dataset, ST-RAP outperforms previous methods, demonstrating the significant benefits of integrating spatial and temporal aspects in real estate appraisal. Our code and dataset are available at 
\todoblue{\url{https://github.com/dojeon-ai/STRAP}}.


\end{abstract}

\maketitle


\section{Introduction} \label{section:intro}

Real estate appraisal, as known as property value estimation, profoundly impacts the financial decisions of individuals and corporations and steers the strategic direction of government policies. Inaccurate appraisals can have severe consequences, leading to substantial financial losses for stakeholders or contributing to government budget deficits.
The 2008 subprime mortgage crisis is a representative example of the potential economic repercussions caused by erroneous real estate appraisal \cite{pavlov2011subprime_real, demyanyk2011understanding_subprime}.

To mitigate such risks, data-driven approaches have been adopted to enhance the precision of real estate price predictions. Traditional methods utilize inherent attributes such as size, age, and location, coupled with simple machine learning (ML) models such as linear regression \cite{csipocs2008linear, ahn2012using_ridge_regression}, support vector machine \cite{smola2004tutorial, lin2011predicting_svr}, or multi-layer neural network \cite{peterson2009neural, law2019satellite, poursaeed2018vision_prediction, wang2021deep, chanasit2021real}.
However, these traditional approaches often fail to reflect complex spatial relationships, such as the interactions between properties within residential communities.

Recent studies have attempted to address this limitation by adopting graph neural networks to model spatial relationships between properties \cite{bin2019peer, zhang2021mugrep, li2022regram}. 
These models represent spatial relationships as a graph, with each node denoting a property. 
For example, in MugRep~\cite{zhang2021mugrep}, nodes are connected based on geographical proximity, incorporating neighborhood properties in their predictions. 
Despite these advancements in spatial modeling, these studies often overlooked the importance of temporal aspects in real estate appraisal. Property prices are dynamic, influenced by market trends, economic policies, and community developments over time. This temporal component is often insufficiently represented \cite{zhang2021mugrep, li2022regram} or entirely disregarded \cite{pagourtzi2003real, peterson2009neural}  in previous studies, which could potentially lead to decreased prediction accuracy.

In response, we introduce ST-RAP, a spatio-temporal framework designed for Real estate APpraisal. ST-RAP constitutes a hierarchical model with a temporal and a spatial model \cite{lee2022draftrec}. First, the temporal model captures the dynamic changes in property values. Subsequently, the spatial model aggregates the temporal model's output, integrating the temporal information from neighboring properties. 
Furthermore, different from previous studies, we include amenities (e.g., schools, hospitals) and transportation stations as nodes and employ a heterogeneous graph network to encapsulate the spatial relationship between properties and these entities.

To verify the effectiveness of ST-RAP, we have manually collected 3.6 million real estate transactions in the Republic of Korea from 2016 to 2020, along with local amenities. 
Through comprehensive experiments, we reveal that ST-RAP outperforms previous methods, with the incorporation of temporal trends in ST-RAP notably contributing to enhanced performance.

\begin{figure*}
  \centering
  \includegraphics[width=0.93\textwidth]{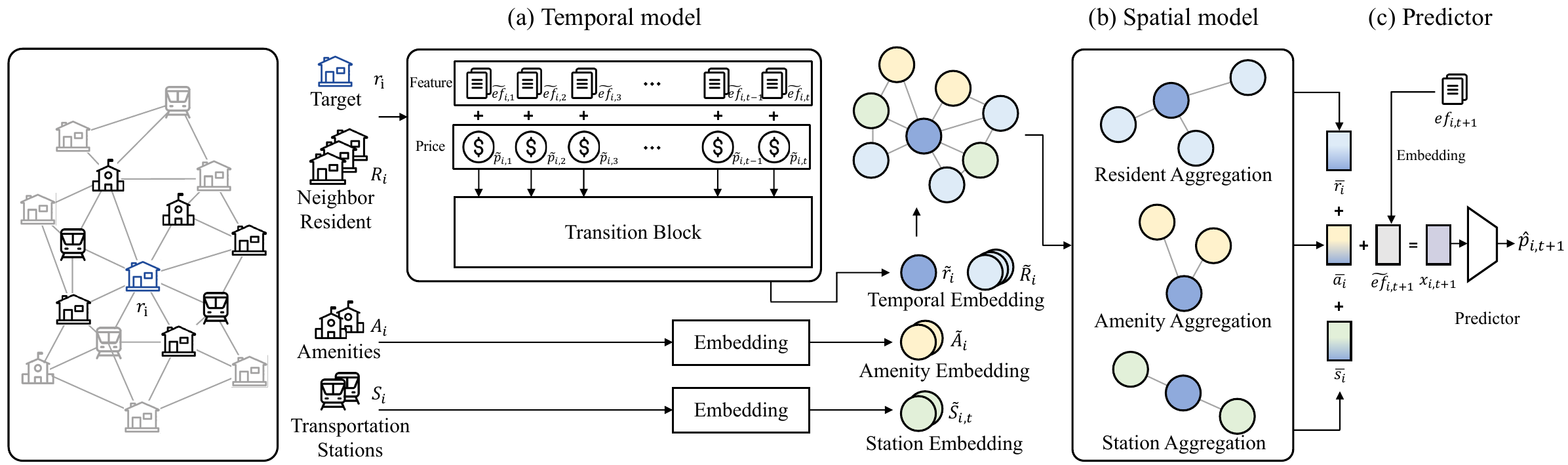}
  \caption{The overall architecture of ST-RAP consists of 3 different models: (a) temporal; (b) spatial, and (c) prediction models.} 
  \label{fig:strap}
\end{figure*}

\section{Related Work}

\noindent
The existing literature on data-driven real estate appraisal largely falls into two categories: (1) traditional automated valuation models (AVMs) and (2) recent graph-based models.
Traditional AVMs focus on property features such as size, age, and location, and employ simple ML models such as linear regression \cite{csipocs2008linear, ahn2012using_ridge_regression}, support vector machines \cite{smola2004tutorial, lin2011predicting_svr}, and neural networks \cite{peterson2009neural, law2019satellite, poursaeed2018vision_prediction, wang2021deep, chanasit2021real}. 
Compared to methods that rely on domain expertise~\cite{Baum2018, guo2014tra, avelino1992, betts2020, shiller1992}, AVMs are efficient and scalable, improving their performance as more data becomes available. 
Despite their advantages, they struggle to capture the spatial
relationships between real estate transactions, which can significantly enhance real estate appraisal.

To address this, recent studies have utilized graph neural networks (GNN) to model spatial relationships between properties \cite{bin2019peer, zhang2021mugrep, li2022regram, xiao2023spatial, xiao2023contextual}. MugRep~\cite{zhang2021mugrep} constructs multiple graphs to capture the relationships between individual transactions and residential communities~\cite{GAT2018}.
ReGram~\cite{li2022regram} further employs an attention module to account for regional differences.


Despite these advancements, prior GNN-based methods face challenges in capturing the \textit{temporal} trends (e.g., price inflation) in real estate values. They attempt to integrate temporal trends by incorporating features such as mean and variance of past transaction values as additional inputs. Nevertheless, such an approach is insufficient for fully capturing the complex, dynamic temporal pattern of real estate values.
Furthermore, these models fail to sufficiently leverage amenities and transportation data, which are critical factors in real estate appraisal.

In response, we introduce ST-RAP, a spatiotemporal framework for real estate appraisal. ST-RAP not only captures temporal dynamics but also includes a heterogeneous GNN to encapsulate the relationships between transactions, and local amenities. 

\section{Preliminaries} \label{section:data}

\subsection{Dataset}

We construct a real estate dataset from the Republic of Korea (2016-2020), which is categorized into four categories: Transaction, Resident, Amenity, and Transportation, as shown in Table \ref{table:dataset}. 
We generate connections between entities where the entities within a 5-kilometer radius are considered to be connected.  

\begin{table}[h]
\centering
\caption{Statistics of real estate dataset.}
\vspace{-2.5mm}
\resizebox{0.47 \textwidth}{!}{
\begin{tabular}{lccc}
\toprule 
Category          & \# of events          & \# of features   & \# of connections to Resident \\
\hline \\[-2ex]
Transaction       & 3,632,022              & 13   & -\\
Resident          & 41,871                 & 25   & 3,449,406     \\
Amenity           & 14,851                 & 10   & 562,521       \\
Transportation    & 13,549                 & 7    & 1,412,919     \\
\bottomrule
\end{tabular}}
\label{table:dataset}
\end{table}

The Transaction category contains 3.6 million records, each defined by 13 attributes including transaction type, date, land area, and price.
The Resident category is composed of 25 features (e.g., location, number of buildings, and the year of construction), with a total of 41,871 records. Each resident is connected based on geographical proximity (total 3.4 million edges).
Additionally, our dataset comprises 15,000 amenities (schools, hospitals, and department stores) and 13,500 transportation stations (train, subway, and bus), generating 1.9 million connections to the residents.


In contrast to previous studies using 0.3 million transactions, our dataset contains 3.6 million transactions that are planned to be publicized.  We believe that our dataset will serve as a valuable resource to promote research in real estate appraisal.

\subsection{Problem Formation}

We represent a residential community as $r_i$, where $1 \leq i \leq N_r$, with $N_r$ being the total number of residents in the dataset.
A transaction event within resident $r_i$ is defined as $e_{i,t}$, with $1 \leq t \leq T_i$ and $T_i$ denotes the total number of transaction events within resident $r_i$.

Each transaction event, $e_{i,t}$, is described by a feature vector $ef_{i,t} \in \mathbb{R}^{d_e}$ and its price $p_{i,t} \in \mathbb{R}$, with $d_e$ representing the dimension of the transaction feature. Hence, we express $e_{i,t}$ as
\begin{equation}
e_{i,t} = \{ef_{i,t}, p_{i,t}\}.
\end{equation}

For each residential community $r_i$, we define a resident feature vector as $rf_i \in \mathbb{R}^{d_r}$ and a sequence of transaction events as $h_{i,1:T_i}$, where $d_r$ denotes the dimension of the resident feature.
Resident $r_i$ is therefore represented as
\begin{align}
r_{i, 1:T_i} &= \{rf_i, h_{i,1:T_i}\} \\
h_{i, 1:T_i} &= [e_{i,1}, e_{i,2}, ..., e_{i,T_i}].
\end{align}
In the sequence of transactions $h_{i, 1:T_i}$, $e_{i, T_i}$ refers to the latest transaction, while $e_{i, 1}$ refers to the earliest transaction in $r_i$.


Residents are connected based on geographical proximity, and we define the set of connected residents to $r_i$ as:
\begin{equation}
\label{eq: connected resident}
    R_{i} = \{r_v \;|\; r_v \text{ is connected to } r_i\}.
\end{equation}

Resident $r_i$ is also connected to nearby amenities and transportation stations.
Each amenity and station is represented as $a_j \in \mathbb{R}^{d_a}$ and $s_j \in \mathbb{R}^{d_s}$, respectively, where $d_a$ and $d_s$ denote the dimensions of the corresponding feature vectors. The sets of connected amenities and transportation stations to $r_i$ are defined as:
\begin{align}
\label{eq: connected amenities}
A_{i} = \{a_v \;|\; a_v \text{ is connected to } r_i\} \\
\label{eq: connected stations}
S_{i} = \{s_v \;|\; s_v \text{ is connected to } r_i\}.
\end{align}

Our goal is to predict the price $p_{i, t+1}$ of the next transaction event $e_{i, t+1}$, given the transaction features $ef_{i, t+1}$.
We seek to incorporate spatiotemporal factors by utilizing the history of past transaction events $r_{i,1:t}$ and connected entities, including residents $R_i$, amenities $A_i$, and transportation stations $S_i$.

We train a function $f$ parameterized by $\theta$ to perform prediction:
\begin{equation}
\hat{p}_{i, t+1} = f_{\theta}(ef_{i, t+1}, r_{i,1:t}, R_i, A_i, S_i).
\end{equation}

\section{Method}

This section presents ST-RAP, a hierarchical architecture with temporal and spatial models. The temporal model captures the temporal trend of transaction prices based on transaction history. Afterward, the spatial model aggregates the output of the temporal model with nearby entities. The overall architecture is illustrated in Figure \ref{fig:strap}.

\subsection{Temporal model}
Here, we describe the temporal model as illustrated in Figure.\ref{fig:strap}(a). 

\vspace{-1mm}
\noindent
\\
\textbf{Embedding.} 
For each transaction event $e_{i,t}$, we linearly embed the price and feature ($p_{i,t}$, $ef_{i,t}$) as $\tilde{p_{i,t}}\in \mathbb{R}^{d_h}$ and $\tilde{ef_{i,t}} \in \mathbb{R}^{d_h}$, where $d_h$ is a hidden dimension of the embedding matrix.
Then, we obtain the transaction embedding, $\tilde{e}_{i,t}$, by summing up these vectors as
\begin{equation}
    \tilde{e}_{i,t} = \tilde{p}_{i,t} + \tilde{ef}_{i,t}.
\end{equation}

Subsequently, for each transaction history $h_{i,1:t}$, the transaction history embedding $\tilde{h}_{i, 1:t}$ is denoted as
\begin{equation}
    \tilde{h}_{i, 1:t} = [\tilde{e}_{i, 1}, \tilde{e}_{i, 2}, ..., \tilde{e}_{i,t}].
\end{equation}
\textbf{Temporal modelling.}
After the embedding stage, the transaction history matrix, $\tilde{h}_{i,1:t}$, is processed by a transition model to effectively capture the temporal dynamics of the transaction events.
Here, we employ a Gated Recurrent Unit (GRU \cite{chung2014empirical}) as the transition model. The output of GRU is defined as 
\begin{equation}
    \tilde{h}'_{i, 1:t} = GRU(\tilde{h}_{i, 1:t}).
\end{equation}

By adding the linearly embedded resident feature, $\tilde{rf}_{i} \in \mathbb{R}^{d_h}$, to the last sequence of the GRU-processed history embedding, $\tilde{h}'_{i,t}$, we derive the resident embedding that reflects temporal information.
\begin{equation}
    \tilde{r}_{i} = \tilde{h}'_{i,t} + \tilde{rf}_{i}.
\end{equation}

\subsection{Spatial model}
Here, we describe the spatial model as demonstrated in Figure.\ref{fig:strap}(b). 

\vspace{-1mm}
\noindent
\\
\textbf{Embedding.}
As described in the equations \ref{eq: connected resident}, \ref{eq: connected amenities}, \ref{eq: connected stations}, the resident $r_i$ is connected to the nearby residents $R_i$, amenities $A_i$, and transportations $S_i$.
Each resident $r_v \in R_i$ is embedded by a temporal model as $\tilde{r}_v \in \mathbb{R}^{d_h}$.
Then, each amenity $a_v \in A_i$ and transportation station in $s_v \in S_i$ are linearly embedded as $\tilde{a}_v \in \mathbb{R}^{d_h}$ and $\tilde{s}_v \in \mathbb{R}^{d_h}$.

\vspace{-1mm}
\noindent
\\ \textbf{Spatial modelling.}
After embedding, we aggregate the features of nearby residents, amenities, and transportation stations to capture the diverse spatial relationships between these entities. 
Given the heterogeneity of the entity types, we utilize a Heterogeneous Graph Neural Network (HGNN \cite{zhang2019heterogeneous}) to construct a comprehensive representation of spatial features.

Given the resident vector $\tilde{r}_i$ and its neighbor entities $\tilde{R}_i, \tilde{A}_i$, and $\tilde{S}_i$, the HGNN is constructed as:
\begin{align}
    \bar{r}_{i} &= W_{rr} (\sigma (W_{rv} \tilde{r}_i + W_{vr} \cdot \text{mean}_{\tilde{r}_v \in \tilde{R}_i} \tilde{r}_v)) + b_{rr} \\
    \bar{a}_{i} &= W_{ar} (\sigma (W_{av} \tilde{r}_i + W_{va} \cdot \text{mean}_{\tilde{a}_v \in \tilde{A}_i} \tilde{a}_v)) + b_{ar} \\
    \bar{s}_{i} &= W_{sr} (\sigma (W_{sv} \tilde{r}_i + W_{vs} \cdot \text{mean}_{\tilde{s}_v \in \tilde{S}_i} \tilde{s}_v)) + b_{sr}
\end{align}
where $\sigma$ denotes the ReLU activation function. The shape of all weight matrices and bias vectors are identical to $W \in \mathbb{R}^{d_h \times d_h}$ and $b \in \mathbb{R}^{d_h}$. Here, the output is defined as $\bar{r}_{i} \in \mathbb{R}^{d_h}, \bar{a}_{i} \in \mathbb{R}^{d_h}, \bar{s}_{i} \in \mathbb{R}^{d_h}$. \\


\noindent \textbf{Prediction.}
Finally, to predict the price of the transaction event $e_{i, {t+1}}$, we construct the state $x_{i,t+1}$ by aggregating the transaction embedding $\tilde{ef}_{i,t+1}$ with the spatiotemporal features $\bar{r}_i, \bar{a}_i, \bar{s}_i$ as
\begin{equation}
    {x}_{i,t+1} = \tilde{ef}_{i,t+1} + \bar{r}_i + \bar{a}_i + \bar{s}_i.
\end{equation}

Then, we predict the price using a feed-forward network as
\begin{equation}
    \hat{p}_{i,t+1} = W_{hp} (\sigma (W_{xh} x_{i,{t+1}} + b_{xh})) + b_{hp}
\end{equation}
where $\sigma$ denotes the ReLU activation, $W_{xh} \in \mathbb{R}^{d_h \times d_h}, W_{hp} \in \mathbb{R}^{d_h \times 1}$ are weight matrices, and $b_{xh} \in \mathbb{R}^{d_h}, b_{hp} \in \mathbb{R}$ are bias terms. \\

\noindent \textbf{Training objective.}
The training objective is to minimize the mean absolute error (MAE) between the predicted price $ \hat{p}_{i,t}$, and the actual price $p_{i,t}$.
We define the loss function $\mathcal{L}$ as the average error across all residents and all transaction events, written as
\begin{equation}
    \mathcal{L} = \frac{1}{N_r T_i}\sum_{i=1}^{N_r}\sum_{t=1}^{T_i}|p_{i,t} - \hat{p}_{i,t}|.
\end{equation}
By optimizing this objective, the model learns to make accurate predictions given the real estate dataset.

\section{Experiments}\label{section:exp}

\subsection{Experimental Setup}
\label{section:experimental_setup}
We split the dataset into training, validation, and test sets in chronological order with a ratio of 90:5:5. 
We trained the model in the training set for 10 epochs, and used the validation set to select the best models. We use Mean Absolute Error (MAE), Root Mean Squared Error (RMSE), and Mean Absolute Percentage Error (MAPE) for evaluation. For MAE and RMSE, the unit price is 1,000,000 KRW.
For all the metrics, the lower score indicates a better performance.

\vspace{-1mm}
\noindent
\\
\textbf{Baselines.} We consider the following baselines: (i) two classical machine learning (ML) baselines (\textbf{L}inear \textbf{R}egression and \textbf{S}upport \textbf{V}ector \textbf{R}egression)~\cite{LR2011}, (ii) a rule-based model \textbf{Repeat} which simply outputs the previous transaction price in the same residential community, (iii) three graph-based models (\textbf{GCN}~\cite{GCN2016}, \textbf{MugRep}~\cite{zhang2021mugrep} and \textbf{ReGram}~\cite{li2022regram}) which captures the spatial dependency between real estate transactions and residential communities.

\vspace{3mm}
\noindent
\textbf{Implementation Details.}
For all the methods, we tuned the hyperparameters using the k-fold cross-validation $(K=5)$. 
We used the AdamW \cite{loshchilov2017decoupled} with varying learning rates of $\{0.1, 0.01, 0.001\}$ and $L_2$ weight regularization strength from $\{0.01, 0.001, 0.0001\}$.
We set the batch size to 128, weight decay to 1e-5, and gradient clipping to 0.5 and trained for 100 epochs.
For MugRep and ReGram, we search the number of attention heads from \{1,2,4,8\}.

\begin{table}[t]
\caption{Performance comparison for real estate appraisal. The results are averaged over 5 random seeds.}
\vspace{-2mm}
\begin{center}
\resizebox{0.42 \textwidth}{!}{
\begin{tabular}{l ccc}
\toprule
Method                        & MAE (M) $\downarrow$    & RMSE (M) $\downarrow$         & MAPE $\downarrow$ \\
\midrule \\[-2.8ex]
LR                            & 105.09 \small{$\pm$1.52} & 194.59 \small{$\pm$1.18}     & 65.62\% \small{$\pm$1.05\%} \\
SVR                           & 90.31 \small{$\pm$3.30} & 137.96 \small{$\pm$1.99}      & 53.22\% \small{$\pm$2.21\%} \\
Repeat                        & 46.01 \small{$\pm$0.00}  & 97.55 \small{$\pm$0.00}      & 28.67\% \small{$\pm$0.00\%}\\
GCN~\cite{GCN2016}            & 50.17 \small{$\pm$2.19}  & 94.13 \small{$\pm$3.45}      & 30.87\% \small{$\pm$0.02\%} \\
MugRep~\cite{zhang2021mugrep} & 35.46 \small{$\pm$2.43}  & 74.08 \small{$\pm$3.64}      & 22.29\% \small{$\pm$0.01\%} \\
ReGram~\cite{li2022regram}    & 34.61 \small{$\pm$4.61}  & 64.48 \small{$\pm$5.03}      & 21.81\% \small{$\pm$0.03\%} \\
\midrule \\[-2.8ex]
ST-RAP    
& \textbf{21.77 \small{$\pm$0.78}}
& \textbf{48.03 \small{$\pm$1.70}}
& \textbf{13.72\% \small{$\pm$0.05\%}}  \\
\bottomrule
\end{tabular}}
\label{table:main_table}
\end{center}
\end{table}

\subsection{Main Results}
\label{section:main_results}

Table \ref{table:main_table} provides the comparative analysis.
Comparing regression models (LR, SVR) to the rule-based model (Repeat), Repeat significantly outperformed regression models.
Repeat can be seen as the simplest version of a temporal model, as it outputs the price of the most recent transaction within a residential community. 
This notable gap implies that the integration of temporal information is crucial in accurately appraising real estate prices.

In addition, we observed that GNN-based models (GCN, MugRep, and Regram), outperform non-spatial models (LR and SVR) across all the evaluation metrics. These findings emphasize the significance of incorporating spatial information for real estate appraisal.

Most importantly, our proposed model, ST-RAP, demonstrates a remarkable improvement over all the comparative models. ST-RAP integrates both temporal and spatial dependencies among residents, amenities, and transportation stations. 
This improvement affirms the effectiveness of reflecting spatiotemporal factors and the use of heterogeneous graphs for an accurate real estate appraisal.

\subsection{Ablation Studies}
\label{section:ablation studies}


\noindent
\textbf{Architecture.} 
We conducted an architecture ablation study by separating the spatial (S-RAP) and temporal (T-RAP) components of ST-RAP. 
The transactions are classified into three groups for this analysis: ALL, which includes all transactions, the COLD group, representing transactions without any historical records within the residents, and the WARM group, representing the remaining transactions with historical records.

\begin{table}[h]
\caption{MAE of S-RAP, T-RAP, and ST-RAP for all transactions (ALL), transactions without any historical records within residents (COLD), and the others (WARM).}
\vspace{-2mm}
\begin{center}
\resizebox{0.37 \textwidth}{!}{
\begin{tabular}{l ccc}
\toprule
Mod  & ALL $\downarrow$   & COLD $\downarrow$ & WARM $\downarrow$ \\
\midrule \\[-2.8ex]
S-RAP                  
& 26.98 \small{$\pm$0.39}  
& 212.20 \small{$\pm$36.86}
& 26.23 \small{$\pm$0.37}  \\
T-RAP                         
& 22.62 \small{$\pm$0.54}
& 230.52 \small{$\pm$12.22}
& 21.99 \small{$\pm$0.54}\\
ST-RAP     
& \textbf{21.77 \small{$\pm$0.78}}
& \textbf{189.42 \small{$\pm$12.68}}
& \textbf{21.21 \small{$\pm$0.76}} \\
\bottomrule
\end{tabular}}
\label{table:srap_vs_trap}
\end{center}
\end{table}
\vspace{-1mm}

Table~\ref{table:srap_vs_trap} shows that ST-RAP outperforms both S-RAP and T-RAP across all metrics, emphasizing the importance of integrating both temporal trends and spatial factors in the model.

\begin{figure}[htbp]
\begin{center}
\includegraphics[width=0.95\linewidth]{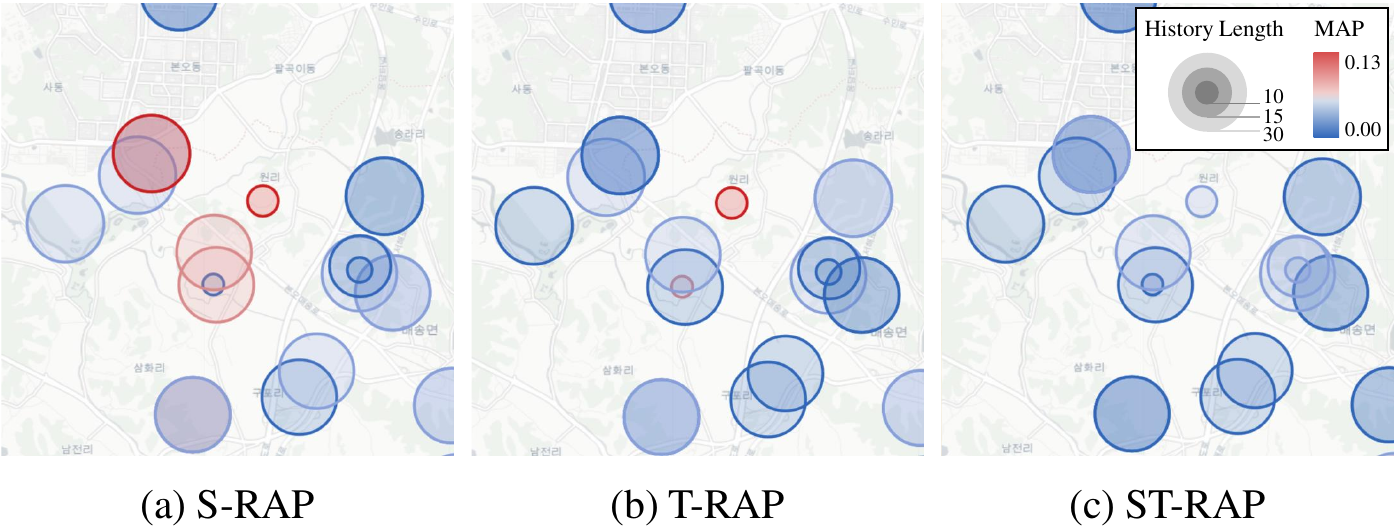}
\end{center}
\vspace{-0.3cm}
\caption{Error distribution of S-RAP, T-RAP, and ST-RAP. The diameter indicates the number of transaction histories.}
\label{fig:srap_trap_strap}
\end{figure}

ST-RAP's improved performance over S-RAP underscores the importance of temporal dynamics. In contrast, the sharp performance drop for T-RAP in the COLD group highlights the importance of spatial modeling when temporal data is scarce. This is further illustrated in Figure \ref{fig:srap_trap_strap}, where S-RAP fails to provide accurate predictions even with an abundance of historical transaction records, and T-RAP's performance diminishes in the absence of such records.

These results reiterate the necessity of incorporating both spatial and temporal aspects for precise real estate appraisal.



\begin{table}[h]
\caption{Ablation of the maximum history length.}
\vspace{-2.5mm}
\begin{center}
\resizebox{0.46 \textwidth}{!}{
\begin{tabular}{l cccccccc}
\toprule
History Len      & 1    & 2   & 5  & 10  & 20  & 30  & 50 \\
\midrule \\[-2.9ex]
MAE (M)               
& 26.98 
& 24.73 
& 22.59
& 22.24 
& 21.93
& \textbf{21.77} 
& 21.89 \\
\bottomrule
\end{tabular}}
\vspace*{-2mm}
\label{table:seq_len}
\end{center}
\end{table}

\noindent
\textbf{History Length.} 
To evaluate the importance of temporal information on transaction value prediction, we conduct an ablation study varying the sequence length (i.e., the number of previous transactions in the residential community). Table~\ref{table:seq_len} displays the performance based on the sequence length for GRU. For both ST-RAP and T-RAP, which consider temporal information, performance improves as the sequence length increases.

\vspace*{-1mm}
\begin{table}[h]
\caption{Ablation of the temporal model.}
\vspace{-2.5mm}
\begin{center}
\resizebox{0.42 \textwidth}{!}{
\begin{tabular}{l ccc}
\toprule
Architecture                 & MAE (M) $\downarrow$      & RMSE (M) $\downarrow$     & MAPE $\downarrow$ \\
\midrule \\[-2.6ex]
Transformer                  
& 24.35 \small{$\pm$0.71}  
& 54.73 \small{$\pm$0.43}  
& 15.32\% \small{$\pm$0.44\%} \\
GRU                          
& \textbf{21.77 \small{$\pm$0.78}}
& \textbf{48.03 \small{$\pm$1.70}}
& \textbf{13.72\% \small{$\pm$0.05\%}}  \\
\bottomrule
\end{tabular}}
\label{table:gru_vs_trm}
\end{center}
\end{table}

\noindent
\textbf{Temporal Model.} While transformer-based models have shown impressive capabilities in handling unstructured time-series data such as language, their performance has been found to be suboptimal when applied to structured time-series data ~\cite{xu2020spatial,cirstea2022transformer,li2020transformertime2,informer2021transformer,wen2023transformerstimeseries,wu2021transformer, zeng2022transformerbad}. Replacing our GRU-based temporal model with a transformer led to a drop in performance, as depicted in Table~\ref{table:gru_vs_trm}. This observation underscores the need for an inductive bias that transformers may lack when dealing with structured temporal trends.

\section{Conclusion} \label{section:conclusion}
We introduced ST-RAP, a novel spatial-temporal framework for real estate appraisal. By integrating the temporal dynamics via a recurrent model and spatial relationships through a heterogeneous graph model, ST-RAP significantly outperforms previous baselines which underscore the importance of reflecting the spatiotemporal factors in real estate appraisal.
In addition, we plan to publicize both our code and dataset to foster future research in this field.


\section{Acknowledgements}
This work was supported by Woomi Construction Co. Ltd. (G01210242), Institute for Information \& communications Technology Planning \& Evaluation(IITP) grant funded by the Korea government(MSIT) (No. 2020-0-00368, A Neural-Symbolic Model for Knowledge Acquisition and Inference Techniques), and the National Research Foundation of Korea (NRF) grant funded by the Korea government (MSIT) (No. NRF-2022R1A2B5B02001913).

\clearpage


\bibliographystyle{ACM-Reference-Format}
\bibliography{reference}
\clearpage



\end{document}